\PassOptionsToPackage{dvipsnames,usenames}{xcolor}
\documentclass[supp]{new_tlp}
\usepackage{supp}
\pubauthor{Jansen}
\usepackage{times}

\usepackage{amsfonts}
\usepackage{amssymb}
\usepackage{amsmath}
\usepackage{xspace}
\usepackage{color}
\usepackage{tikz}
\usetikzlibrary{shapes,arrows}
\usepackage{xspace}
\usepackage[official]{eurosym}
\usepackage{enumitem}
\usepackage{subfigure}

\usepackage[english]{babel}
\usepackage{import}

\usepackage{ifthen}
\usepackage{url}
\usepackage{amssymb}
\usepackage{amsmath}
\usepackage{import}
\usepackage{amsmath}

\usepackage[usenames,dvipsnames]{xcolor}
\usepackage{xspace}
\usepackage{datatool}
\usepackage{glossaries}

\newcommand{\m}[1]{\ensuremath{#1}\xspace}

	\newcommand{\lrule}{\m{\leftarrow}}
	\newcommand{\cause}{\m{\stackrel{c}{\lrule}}}

	\newcommand{\struct}{\m{I}}

	\usepackage{xparse}
	\NewDocumentCommand\inter{g+g}{%
	  \IfNoValueTF{#1}
	    {\struct}
	    {\m{#1^{#2}}}}

	\renewcommand{\int}{\m{\mathbb{Z}}}

	\NewDocumentCommand\subs{g+g}{%
	  \IfNoValueTF{#1}
	    {\m{/}}
	    {\m{#1/ #2}}}

	\newcommand{\logicname}[1]{\text{\sc #1}\xspace}

	\newcommand{\idp}{\logicname{IDP}}

	\newcommand{\idpthree}{\logicname{IDP$^3$}}

	\newcommand{\minisatid}{\logicname{MiniSAT(ID)}}

	\newcommand{\fodotidp}{\logicname{FO(\ensuremath{\cdot})\ensuremath{^{\mathtt{IDP}}}}}
	
	\newcommand{\fodot}{\logicname{FO(\ensuremath{\cdot})}}

\newcommand{\ouracronym}[3]{%
	\newacronym{#1}{#2}{#3}
	\expandafter\newcommand\csname #1\endcsname{\gls{#1}\xspace}%
}
	\ouracronym{FO}{FO}{first-order logic}
	\ouracronym{MX}{MX}{Model Expansion}
	\ouracronym{MO}{MO}{Model Optimization}
	\ouracronym{ASP}{ASP}{Answer Set Programming}
	\ouracronym{TP}{TP}{Theorem Proving}
	\ouracronym{LP}{LP}{Logic Programming}
	\ouracronym{CP}{CP}{Constraint Programming}
	\ouracronym{KR}{KR}{Knowledge Representation}
	\ouracronym{CSP}{CSP}{Constraint Satisfaction Problem}
	\ouracronym{SMT}{SMT}{SAT Modulo Theories}
	\ouracronym{KBS}{KBS}{knowledge-base system}
	\ouracronym{NNF}{NNF}{Negation Normal Form}
	\ouracronym{FNNF}{FNNF}{Flat Negation Normal Form}
	\ouracronym{DefNNF}{DefNNF}{Definition Negation Normal Form}
	\ouracronym{CDCL}{CDCL}{Conflict-Driven Clause-Learning}
	\ouracronym{LCG}{LCG}{Lazy Clause Generation}

	\makeatletter
	\def\ifenv#1{
	\def\@tempa{#1}%
	\def\@ttempa{#1*}%
	\ifx\@tempa\@currenvir
	\expandafter\@firstoftwo
	\else
	\expandafter\@secondoftwo
	\fi
	}
	\makeatother

	\newcommand{\ddrule}[4]{\ensuremath{#1 \leftarrow #2 & \{#3\} & #4}}
	\newcommand{\drule}[2]{\ensuremath{#1 & \leftarrow & #2}}

	\newcommand{\darule}[4]{\ensuremath{#1 \leftarrow #2 & \{#3\} & #4}}
	\newcommand{\arule}[2]{\ensuremath{#1 \, &\leftarrow \, #2}}

	\newcommand{\LNDRule}[2]{
	\ifenv{array}
	{\drule{#1}{#2}}
	{ \ifenv{align}
		{\arule{#1}{#2}}
		{\ifenv{align*}
		{\arule{#1}{#2}}
		{ERROR: using LDRule in unsupported environment: \@currenvir}
		}
	}
	}

	\newcommand{\LDRule}[4]{
	\ifenv{array}
	{\ddrule{#1}{#2}{#3}{#4}}
	{ \ifenv{align}
		{\darule{#1}{#2}{#3}{#4}}
		{\ifenv{align*}
		{\darule{#1}{#2}{#3}{#4}}
		{ERROR: using LDRule in unsupported environment: \@currenvir}
		}
	}
	}

	\NewDocumentCommand\LRule{m+g+g+g}{%
		\IfNoValueTF{#2}%
		{#1.&}{%
		\IfNoValueTF{#3}
		{\LNDRule{#1}{#2}}
		{\LDRule{#1}{#2}{#3}{#4}}%
		}
	}

	\NewDocumentCommand\CLRule{m+g}{%
	\ifenv{array}
	{\cdrule{#1}{#2}}
	{ \ifenv{align}
		{\carule{#1}{#2}}
		{\ifenv{align*}
			{\carule{#1}{#2}}
			{ERROR: using CLRule in unsupported environment: \@currenvir}
		}
	}
	}

	\NewDocumentCommand\carule{m+g}{%
		\IfNoValueTF{#2}
			{\ensuremath{#1.}}
			{\ensuremath{#1 \, &\cause \, #2}}}
	\NewDocumentCommand\cdrule{m+g}{%
		\IfNoValueTF{#2}
			{\ensuremath{#1.}}
			{\ensuremath{#1 & \cause & #2}}}

	\newcommand{\algrule}[4]{
	\hbox{{#1}:}& 
	\quad #2 ~\longrightarrow~ #3 
	\hbox{~ if } #4\\
	}

	\newcommand{\AlgoRule}[4]{
	\ifenv{array}
	{\algrule{#1}{#2}{#3}{#4}}
		{ERROR: using AlgoRule in unsupported environment: \@currenvir}
	}

\newcommand{\commentstyle}{\color{Gray}}
	\usepackage{listings}

	\lstdefinelanguage{idp}{
		morekeywords=[1]{namespace,vocabulary,theory,structure,procedure,term,set,formula, spec, specification},
		morekeywords=[2]{include,using,type,isa,contains,partial,extern,LFD,GFD,constructed,from,constraint,func,pred,supertype,of,subtype,define},
		morekeywords=[3]{int,float,char,string,nat},
		morekeywords=[4]{if,then,else,for,end},
		morecomment=[s]{/*}{*/},	
		morecomment=[l]{//}
	}
	\newcommand{\ignore}[1]{}

	\newboolean{nocomments}
	\setboolean{nocomments}{false}

	\newboolean{commentmargin}
	\setboolean{commentmargin}{true}

	\newcommand{\namedcomment}[3]{
		\ifthenelse{\boolean{nocomments}}
		{} %IF no comments, write nothing
		{ %Otherwise
			\ifthenelse{\boolean{commentmargin}}
				{ {\color{#3} \marginpar{\color{#3}\sc #2}#1}  } %Name in margin
				{  {\color{#3} {\sc #2}: #1}  } %Name not in margin
		}
	}
	\newcommand{\mnamedcomment}[3]{\ifthenelse{\boolean{nocomments}}{}{{\marginpar{ \color{#3}{\sc #2}:#1}}}}

	\usepackage{soul}

\newcommand{\keyword}[2]{%
	\expandafter\newcommand\csname #1\endcsname{#2\xspace}%
	\expandafter\newcommand\csname #1s\endcsname{#2s\xspace}%
	\expandafter\newcommand\csname #1ness\endcsname{#2ness\xspace}%
}

\newcommand{\figref}[1]{Figure~\ref{#1}}

\title{Model revision inference for extensions of first order logic}
\author{Joachim Jansen}
\pdfoutput=1

\begin{document}
\maketitle
% \tableofcontents

\begin{abstract}
I am Joachim Jansen and this is my research summary, part of my application to the Doctoral Consortium at ICLP`14.
I am a PhD student in the Knowledge Representation and Reasoning (KRR) research group, a subgroup of the Declarative Languages and Airtificial Intelligence (DTAI) group at the department of Computer Science at KU Leuven.
I started my PhD in September 2012.
My promotor is prof. dr. ir. Gerda Janssens and my co-promotor is prof. dr. Marc Denecker.

I can be contacted at {\tt joachim.jansen@cs.kuleuven.be} or at:
\begin{verbatim}
Room 01.167
Celestijnenlaan 200A 
3001 Heverlee 
Belgium
\end{verbatim}
An extended abstract / full version of a paper accepted to be presented at the Doctoral Consortium of the 30th International Conference on Logic Programming (ICLP 2014), July 19-22, Vienna, Austria
\end{abstract}

\section{Problem description}
\subsection{Introduction}
% Het \idp systeem is een state-of-the-art systeem voor declaratief probleemoplossen; complexe zoek- en optimalisatieproblemen worden op een effici\"ente en generieke manier opgelost.
% Wanneer door omstandigheden een gevonden oplossing lichtjes wordt aangepast, of indien we bijkomende informatie krijgen, is het vaak wenselijk om op een goedkope manier een {\em revisie} voor de huidige oplossing te maken: we wensen ze aan te passen aan deze nieuwe informatie.
% Men wil dan vertrekken van de oude (bijna)-oplossing en mits enkele kleine wijzigingen deze transformeren tot een oplossing waarin de bijkomende informatie verwerkt is.
% Er bestaat momenteel geen enkele effici\"ente en algemene oplossing voor dergelijke problemen; de enige manier waarop dit momenteel wordt opgelost is door special-purpose algoritmes te schrijven. 
% Met dit project willen we dit soort revisie problemen op een algemene manier oplossen in de context van een expressieve modelleertaal.

The \idp system is a state-of-the-art system for declarative problem solving; complex search- and optimizationproblems are solved in an efficient and generic manner.
As time passes on however, the found solution has to be {\it revised}: new information (e.g., changed circumstances) has to be taken into account.
In this case it is desirable to start from the old (near-)solution and by perfroming a limited amount of changes transform it into a solution in which the new information is processed. %processed?
At the moment there are no efficient, general solutions for these kind of problems; the only way this problem is currently solved is by writing special-purpose algorithms.
During my thesis I would like to devise a general way to solve these problems using \idp as a system supporting an expressive modeling language.

% Het {\em Knowledge Base System} (KBS)~\cite{IDPKBS} paradigma is een declaratieve aanpak waar men specifieert {\em wat} er opgelost moet worden, in plaats van procedures te schrijven die zeggen {\em hoe} dit te doen~\cite{Apt03,synthesis/2012Gebser}.
% % Dit resulteert in duidelijkere softwareoplossingen, waarbij de hoeveelheid code met 90\% kan verminderen~\cite{iclp/Blockeeletall12}.
% Een KBS stelt de kennis in een expliciete vorm voor met een expressieve modelleertaal en voorziet hiervoor inferenties om verschillende soorten problemen op te lossen.
% De expressieve modelleertaal heeft als voordeel dat domeinen met erg complexe of snel veranderende kennis op een korte, overzichtelijke manier uitgedrukt kunnen worden.
% Bovendien kan kennis binnen hetzelfde toepassingsgebied herbruikt worden om verschillende problemen in deze context op te lossen.
% Omdat de inferenties domeinonafhankelijk zijn, kunnen ook deze herbruikt worden over verschillende probleemdomeinen heen (zie sectie~\ref{sec:toepassingen}).

The {\em Knowledge Base System} (KBS)~\cite{IDPKBS} paradigm is a declarative approach in which one specifies {\em what} needs to be solved, instead of writing procedures that depict {\em how} to do this~\cite{Apt03,synthesis/2012Gebser}.
A KBS represents the knowledge in its explicit form using an expressive modeling language and provides inferences to solve different kinds of problems.
The expressive modeling language has as advantage that domains with a very complex or quickly changing knowledge can be expressed in a concise and clear way.
Additionally, knowledge can be reused to solve different problems sharing the same scope.
Because the inferences are domain independent, they can be reused across different scopes as well.

% \'E\'en van deze belangrijke inferenties is modelrevisie; het aanpassen van een bestaande oplossing aan nieuwe vereisten.
% In treinherplanning komen er bijvoorbeeld onvoorziene omstandigheden voor (spoordefecten, vertragingen, kabeldiefstallen) en moet de planning aangepast worden aan nieuwe vereisten.
% Deze inferentie probeert zoveel mogelijk van de oorspronkelijke oplossing (treinplanning) te behouden bij het verwerken van de aanpassing.
% zich schikken naar de nieuwe situatie.
% Het is namelijk effici\"enter om dit probleem op te lossen door een beperkt aantal aanpassingen te maken aan de huidige planning.
% Je wil immers dat als een trein vertraging oploopt in Brugge dit probleem lokaal opgelost wordt en niet opnieuw een planning opstellen voor alle treinen in Belgi\"e.

One of these inferences is model revision; the adaptation of an existing solution to new information.
In a train dispatching toy-problem for example there are a plethora of unforeseen circumstances (e.g., track defects, delays, copper cable thefts) and the dispatching schedule needs to be adapted to new requirements.
Model revision also tries to maintain as much as possible of the original solution (dispatching schedule) when processing the change.
This is a consequence of the solution technique that is generally efficient (start from the `old' solution and apply a limited amount of changes), but is also a desirable property of the computed solution.
Indeed, when a train is delayed in Paris, it doesn't make a lot of sense to change the dispatching schedule of trains in London when this is not necessary.

% Modelrevisie laat ons toe {\em flexibel} om te gaan met een berekende oplossing (uurrooster/treinplanning/netwerkconfiguratie) door ons nieuwe vereisten te laten opleggen (iemand is ziek/een trein heeft vertraging/een server valt uit) aan een bestaande oplossing.
% Hoewel dit flexibel omgaan met een oplossing essentieel is voor een volwaardig KBS, is er nog geen ondersteuning voor modelrevisie in de context van een expressieve modelleertaal en staat het onderzoek naar modelrevisie nog in de kinderschoenen.
% Soortgelijk onderzoek gebeurde wel in technieken zoals incremental constraint programming~\cite{cacm/Freeman-BensonMB90} en reactive answer set programming~\cite{lpnmr/GebserTRT}.
% Hierin onderzoekt men echter maar een beperkte vorm van aanpassingen: er wordt rekening gehouden met bepaalde types van nieuwe kennis, maar er is geen mogelijkheid algemene, eerder onvoorziene aanpassingen te maken.
% Voor eerste orde logica is er een basisalgoritme dat wel rekening houdt met algemene aanpassingen~\cite{inap/WittocxDD09}.

\subsection{Model revision: a motivating example}
\label{sec:mr_example}

\begin{figure}
\begin{center}
\label{fig:exampleStructureGraph}
 \includegraphics[width=.6\textwidth,bb=0 0 1149 534]{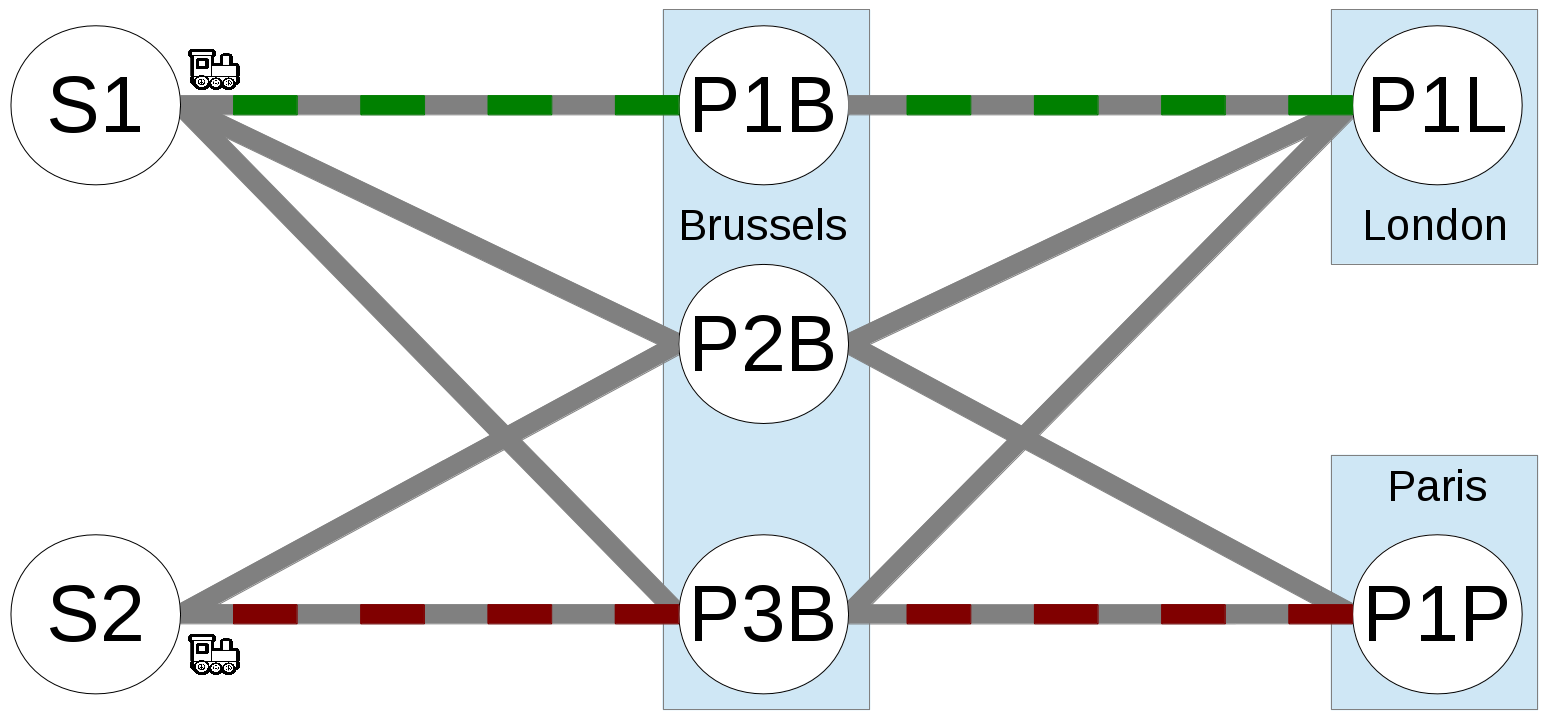}
 % example_structure.png: 1548x719 pixel, 97dpi, 40.53x18.83 cm, bb=0 0 1149 534
\caption{An example of a train routing situation}
\end{center}

\end{figure}

\begin{figure}[h]\centering
%   \begin{subfigure}[l]{0.45\textwidth}
  \subfigure[A high-quality revised model]{
    \includegraphics[width=0.45\textwidth,bb=0 0 1079 503]{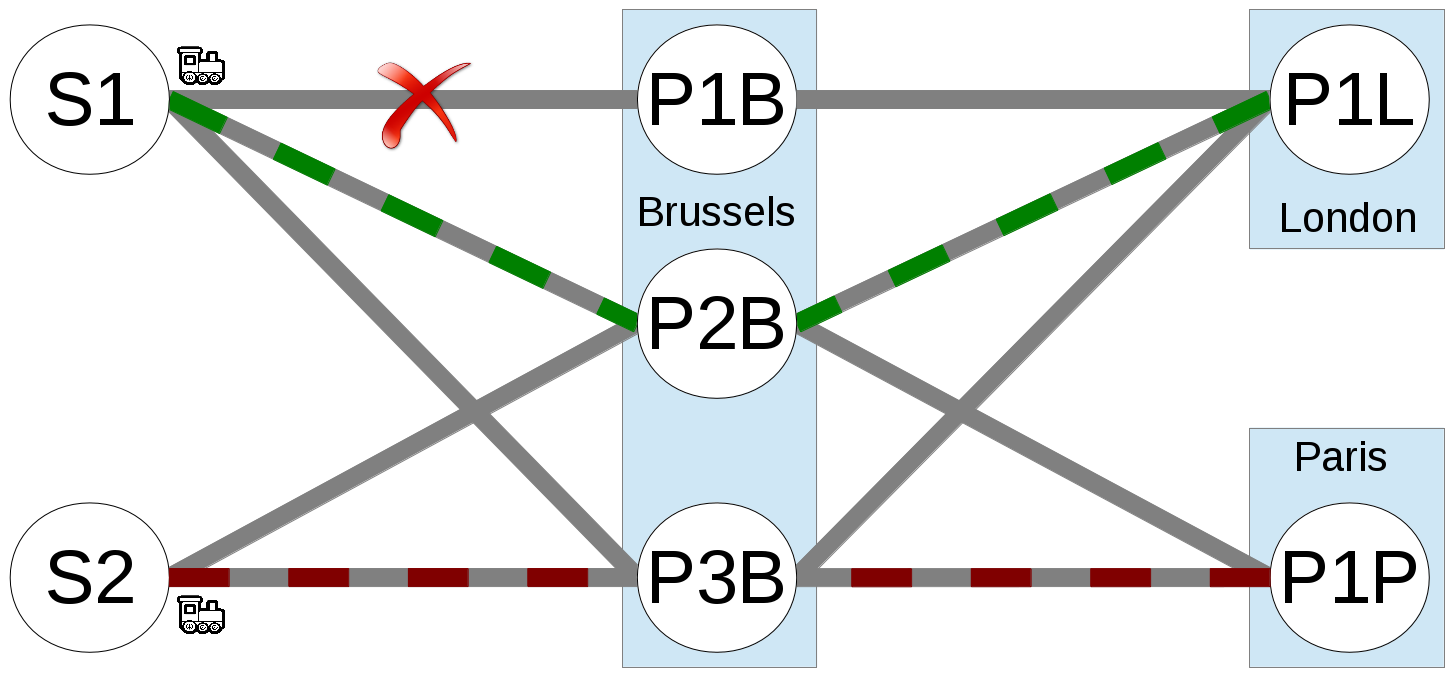}
  % mr_graph_good_alt.png: 935x432 pixel, 96dpi, 24.74x11.43 cm, bb=0 0 1079 503
%     \caption{Een goed gereviseerd model}
    \label{fig:mr_graph_good_alt}
  }
%   \end{subfigure}
% ~
%   \begin{subfigure}[r]{0.45\textwidth}
  \subfigure[A low-quality revised model]{
    \includegraphics[width=0.45\textwidth,bb=0 0 969 452]{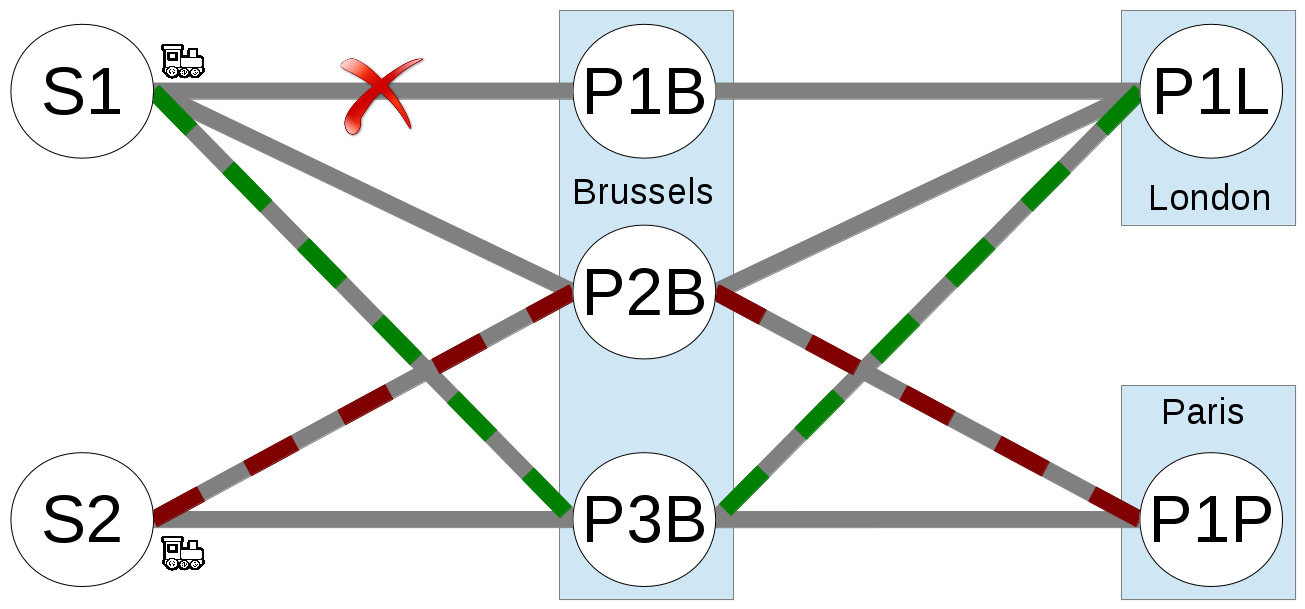}
    % mr_graph_bad_alt.png: 929x429 pixel, 96dpi, 24.58x11.35 cm, bb=0 0 969 452
%     \caption{Een slecht gereviseerd model}
    \label{fig:mr_graph_bad_alt}
  }
%   \end{subfigure}
\caption{Examples of high- and low-quality revised models}
\label{fig:mr_graph_revised}
\end{figure}

Here we introduce a small motivating example of the model revision inference using the situation depicted in \figref{fig:exampleStructureGraph}.
In this figure the train tracks are indicated using grey lines between nodes.
The train starting in S1 (which we will also call Train1) has to go past stations Brussels and London and the train that starts in S2 ((which we will call Train2) has to visit stations Brussels and Paris.
The dispatched route for this is indicated using a green dotted line, that of Train2 is indicated with a red dotted line.
Imagine the train track between S1 (Shunting 1, shuntings are intermediary crossroads in train tracks where one can change direction) and P1B (Platform 1 in Brussels) is detected to have broken down.
By using model revision we can construct a new route for Train2 in S1 that does not use any broken down train tracks.
% \begin{figure}\centering
%  \includegraphics[width=.8\textwidth]{./mr_graph.png}
%  % mr_graph.png: 925x426 pixel, 96dpi, 24.47x11.27 cm, bb=0 0 694 319
% \caption{Een voorbeeld van een bestaande treinplanning waar iets misgaat: het spoor tussen W1 en P1A wordt onbruikbaar}
% \label{fig:mr_example}
% \end{figure}
\figref{fig:mr_graph_good_alt} shows a high-quality revised model: a route has been found for Train1 without changing too much to the existing dispatching.
\figref{fig:mr_graph_bad_alt} shows a low-quality revised model: the route for Train1 is correct but an unnecessary change to the route of Train2 was made.
The change to the route of Train2 was needed because there is a requirement that states that two trains cannot enter the same station on the same platform (at the same time).

\subsection{Formal definition of model revision}
\label{sec:mr_formal_definition}

The formal definition for the model revision problem is as follows~\cite{inap/WittocxDD09}:

Given a \fodot theory $T$, a model $M$ for this theory and a collection of domain atoms $C$.
Henceforth $C$ are called the {\em required changes}.
In the example $C$ is the usage of the tracks between S1 and P1B.
Solving {\em model revision} for $\langle T, M, C \rangle$ means searching a new model $M'$ of $T$ such that all domain atoms in $C$ all have a different value compaired to their old one in $M$.
$M'$ is also called the {\em revised model}.
\figref{fig:mr_graph_revised} shows two possible revised models for the example; the broken down train tracks are not used in either case.

In addition to the required changes, one usually has to change other parts of the original solution as well to construct the revised model.
We call these other changes between $M$ and $M'$ the {\em additional changes} and denote them with $S$.
In the example the usage of the new route is the additional change.

Often it is not desirable that the entire original model $M$ is be changed; some elements are {\em immutable}.
In the example the structure of the train tracks is considered immutable: we are not interested in new solutions that would require us to build additional train tracks (e.g., one between London and Paris).
These immutable elements in the problem domain are represented by the {\em limitation} $G$, a set of domain atoms whose value must remain fixed.
The revision problem for $\langle T, M, C, G \rangle$ is the same as the revision problem for $\langle T, M, C \rangle$, except for the extra condition that the additional changes cannot include any of the limitations (i.e., $S$ is disjoint with $G$).

\section{Existing Literature}

Model revision allows us to flexibly use with a computed solution by imposing new restrictions.
Although this kind of flexible reasoning is essential to a KBS, there is no research for model revision (in its general sense) in the context of an expressive modeling language.
Comparable research has been performed in areas of incremental constraint programming~\cite{cacm/Freeman-BensonMB90} and reactive answer set programming~\cite{lpnmr/GebserTRT}.
In this research only a limited form of new requirements are supported: one takes into account specific forms of new types of knowledge, but e.g. there is no way to apply previously onforeseen changes.
Recent research are also trying to tackle this problem on the SAT level~\cite{lncs/abio}.
These SAT-level techniques are interesting for the implementation that will be provided eventually because \idp uses a SAT solver in its workflow, but do not work in the context of a complex modeling language.
There has also been work on trying to construct the solution in such a way that it is `robust' w.r.t. changes~\cite{journals/jair/robustnessInCP}.
For first order logic there is a basic algorithm that takes general changes into account~\cite{inap/WittocxDD09}.
This will serve as a starting point for my thesis.

\section{Background}

This section contains a short introduction to the used terminology.
The following concepts are introduced briefly: Knowledge Base System paradigm~\cite{IDPKBS}, \fodot~\cite{corr/Blockeel13}, and the \idp system~\cite{DeCatGTTV2013,url:idp}.

\subsection{The language: \fodot}

Each declarative system requires a language in which the problems are represented.
This language is preferably expressive, so the problem domain can be intuitively expressed.
The \fodot family of languages has been developed at the KRR group for this purpose.

\fodot is a family of expressive knowledge representation languages that extend classical First Order Logic (FO) with various concepts.
Apart from the logical symbols ($\wedge$, $\vee$, $\neg$, $\Rightarrow$, $\Leftrightarrow$, $\exists$, $\forall$), \fodot also contains:

\begin{description}
 \item [Inductive definitions] are represented as a set of defining rules.
 \item [Set expressions] of the form $\{ x~y : p(x) \wedge q(y) \wedge r(x,y)\}$ represent the set of all combinations of $x$ and $y$ such that $p(x) \wedge q(y) \wedge r(x,y)$. 
 \item [Aggregates] express the result of an aggregate function of a set expression together with a cost function (for each element in the set).
 The following aggregate functions are supported: minimum, maximum, sum, product and cardinality.
 
 \item [Expressive quantifiers] such as $\exists_{=1}$ (there exists exactly one), $\exists_{\geq 2}$ (there exists at least two) ...
 
 \item [Types and subtypes]: each variable is typed in \fodot.
 
 \item [(Partial) functions]. These are non-Herbrand functions.
 
 \item [Arithmetic operators] such as $+$ $-$, $\times$, $\div$, $| x |$, and $\%$.
\end{description}

A problem specification in \fodot consists of at least three parts: a {\em vocabulary} that depicts the domain ontology, a {\em theory} containing the constraints for this problem, and a {\em structure} that contains the known data about the problem.

For a more hands-on introduction to \fodot and \idp, the reader is directed to our webpage of examples at \url{http://dtai.cs.kuleuven.be/krr/software/idp-examples}

\subsection{Knowledge Base System}

\begin{figure}\centering
 \includegraphics[width=.8\textwidth,bb=0 0 1159 274]{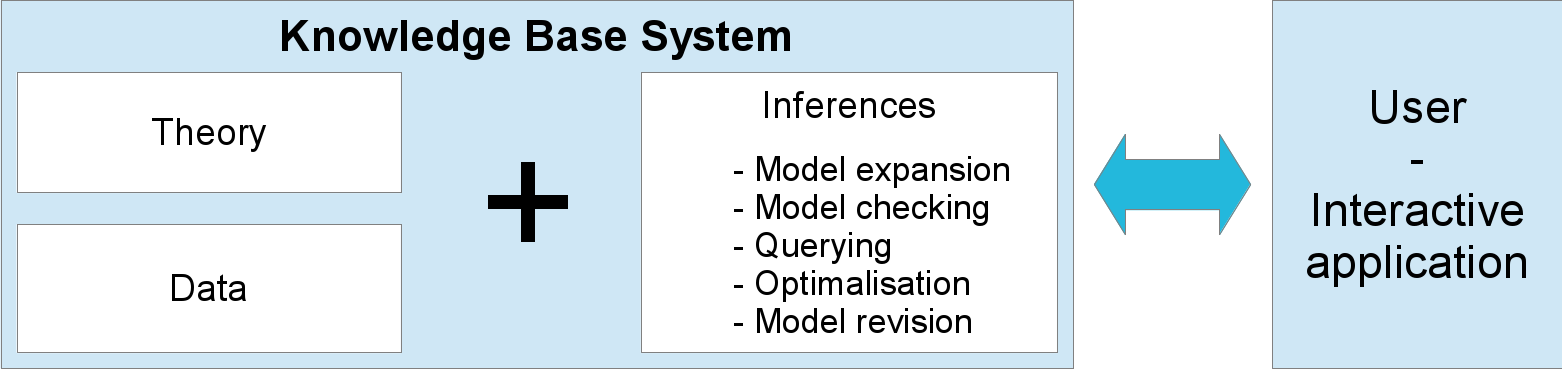}
 % kbs.png: 885x575 pixel, 96dpi, 23.41x15.21 cm, bb=0 0 1159 274
\caption{A conceptual representation of a Knowledge Base System}
\label{fig:kbs}
\end{figure}

In a Knowledge Base System (KBS) the data and knowledge (expressed in the modeling language, e.g., \fodot) are maintained in a {\em Knowledge Base}.
A KBS then offers a variety of inferences to solve problems with the knowledge.
A conceptual representation of a KBS is displayed in \figref{fig:kbs}.

Among these inferences are {\em model expansion} (extend a three-valued structure such that it satisfies a theory), {\em model checking} (verify whether a given structure satisfies a theory), {\em optimization} (extend a three-valued structure to a two-valued structure that satisfies a theory that has the least cost), and {\em model revision} (see Section~\ref{sec:mr_formal_definition}).

\subsection{The \idp system}

The \idp system is a state-of-the-art implementation of the KBS paradigm using \fodotidp as its modeling language.
The workflow of the \idp system is as follows~\cite{DeCatGTTV2013}.
First the \fodot theory is ground into a low-level propositional representation.
This representation is called ``Extended CNF'' or ECNF.
It is an extension of CNF with concepts such as inductive definitions (that are ground).
Next \idp uses a SAT-solver, \minisatid, to generate solutions based on the grounding.
\idp as well as \minisatid are open-source and available at \url{https://bitbucket.org/krr/idp} and respectively \url{https://bitbucket.org/krr/minisatid}.

The goal of my thesis is to provide support for model revision in the \idp system.

\section{Goal of the research}
 
The goal of my PhD thesis is to develop logic inference methods for different forms of model revision in the context of the \fodot modeling language.

In order for this to be possible, we need a mechanism to reason about changes propagating through a theory.
To this end, the {\em approximating definition} for a theory~\cite{phd/Wittocx10,phd/Vlaeminck12,INAP/VaezipoorMM11} needs to be computed and used to propagate impact of a change to the solution throughout the theory of the problem.
The theory behind this currently supports basic FO.
For my thesis, I will extend the scope of approximating definition to theories containing more expressive constructs such as inductive definitions, aggregates...

Because the approximating definition is a definition that needs to be calculated, there need to be efficient techniques for doing so.
It was proposed in \cite{phd/Wittocx10} that the definition can be evaluated using any external system that can evaluate definitions (or rules).

For model revision there are typically a multitude of possible revisions.
There is a need for proper {\em criteria} that quantify the quality of a revision.
I intend to construct criteria using a {\em domain independent} as well as a {\em domain dependent} approach.
For the domain independent criteria some brute-force metrics such the number of changed domain atoms will be used.
In order to properly support domain dependent criteria, a user needs to able to express which revisions are preferred over others. 
This can be done either by expressing them beforehand using some sort of cost function.
For this the knowledge representation language needs to be extended.
Another way to do this is to let the user interactively guide the search process for the revision, indicating which choices are preferred.

\section{Current status of the research}
The first part of my PhD consisted of constructing an interface between {\sc XSB} and \idp for calculating definitions that can be completely evaluated.
For this work, the inductive definitions are transformed into rules for tabled Prolog.
This was published in TPLP~\cite{tplp/Jansen13}.

Further I extended \idp to compute the approximating definition using the existing theory concerning this topic.
Additionally, \idp was also extended with the possibility to making the input structure as two-valued as possible before grounding using the approximating definition~\cite{DeCatGTTV2013} as an alternative approach to the ``Ground With Bounds'' (GWB) technique depicted in~\cite{phd/Wittocx10,phd/Vlaeminck12}.
Currently benchmarks are being run to compare the two approaches.
According to~\cite{INAP/VaezipoorMM11} the new approach using approximating definition outperforms the classical GWB technique because it will always compute all possible unit propagation possible (at SAT-level) beforehand. GWB on the other hand sometimes performs cutoffs to increase performance.
Preliminary results however contratict this claim.

Another claim from~\cite{INAP/VaezipoorMM11} is currently being investigated: a ``smarter'' grounding will affect the search tree as well.
A smarter grounding can contain fewer introduced symbols (i.e., Tseitins) because it was detected beforehand that they need not be generated at all.
Since these Tseitins are not removed by performing unit propagation at SAT-level, a smarter grounding thus contains (according to the above authors) possbily fewer ``autarkies'' - irrelevant parts of the search space in which the solver possbibly can waste time.
Currently experiments are being run that compare the search behaviour of solver runs on smart, respectively ``naive'' groundings.

\section{Preliminary results}
Benchmarks over problems in the $P$ complexity class that are generally solved by evaluating definitions for completely given structures show that a great speedup is achieved compared to the classical approach~\cite{tplp/Jansen13}.

Preliminary results (a complete study is being performed) suggest that making the input structure as two-valued as possible before grounding using approximating definitions is not superior to its counterpart the classic GWB workflow already implemented in \idp.
Additionally, there were only very few problems where the grounding was smaller.

\section{Open Issues}
Tasks that still need addressing are the extension of the approximating definition for theories that contain more expressive constructs such as inductive definitions, aggregates...
Additionally, the solver \minisatid will need to be adapted to support model revision.
For support of interactively searching for a revision, the solver workflow also needs to be updated to work interactively with user input.

\newpage
\bibliographystyle{acmtrans}

\end{document}